\documentclass[runningheads]{llncs}

\usepackage{eccv}

\usepackage{eccvabbrv}
\usepackage{graphicx}
\usepackage{booktabs}

\usepackage[accsupp]{axessibility}  %

\usepackage{hyperref}

\usepackage{orcidlink}
\usepackage{float}
\usepackage{tabularx}

\begin{document}

\title{ArCSEM: Artistic Colorization of SEM Images via Gaussian Splatting}

\titlerunning{ArCSEM}

\author{Takuma Nishimura\inst{1} \and Andreea Dogaru\inst{1} \and \\Martin Oeggerli\inst{2} \and Bernhard Egger\inst{1}}
\authorrunning{T.~Nishimura et al.}

\institute{
Friedrich-Alexander-Universität Erlangen-Nürnberg\\
\email{\{takuma.nishimura, andreea.dogaru, bernhard.egger\}@fau.de}
\and
Micronaut\\
\email{info@micronaut.ch}
}
\maketitle

\begin{abstract}
Scanning Electron Microscopes (SEMs) are widely renowned for their ability to analyze the surface structures of microscopic objects, offering the capability to capture highly detailed, yet only grayscale, images. To create more expressive and realistic illustrations, these images are typically manually colorized by an artist with the support of image editing software. This task becomes highly laborious when multiple images of a scanned object require colorization. We propose facilitating this process by using the underlying 3D structure of the microscopic scene to propagate the color information to all the captured images, from as little as one colorized view. We explore several scene representation techniques and achieve high-quality colorized novel view synthesis of a SEM scene. In contrast to prior work, there is no manual intervention or labelling involved in obtaining the 3D representation. This enables an artist to color a single or few views of a sequence and automatically retrieve a fully colored scene or video. Project page: \url{https://ronly2460.github.io/ArCSEM}

\end{abstract}

\section{Introduction}
\label{sec:intro}
Throughout history, different art forms have been used to express ones' creative perspective and share it with the world. As the society and technology evolve, artistic endeavours reach beyond set boundaries in the exploration of imaginative challenges. Visual arts in particular are being revitalized by novel approaches at the intersection of computer graphics, computer vision and human creativity, where artistic expression entangles with academic research. 

Following the advancement of deep learning-based methods, applications such as neural style transfer \cite{gatys2016image, huang2017arbitrary, johnson2016perceptual, zhu2017unpaired}, image colorization \cite{he2018deep, zhang2016colorful, cheng2015deep}, text-prompted image generation \cite{rombach2022high, zhang2023adding} and editing \cite{brooks2023instructpix2pix, hertz2022prompt} have become easily-accessible for both professional artists and curious creators. Though these methods are capable of impressive results, they are limited to the 2D domain. Another research area in computer vision that has seen tremendous improvements in the recent years is optimizing 3D scenes from multi-view inputs. Either with the goal of synthesising novel views or reconstructing the underlying 3D surface, the scene representation plays an important role in achieving high-quality results. Popular choices are point clouds \cite{rakhimov2022npbg++, ruckert2022adop, franke2024trips}, neural fields \cite{mildenhall2021nerf, wang2021neus, barron2021mip}, voxels \cite{fridovich2022plenoxels}, hybrid representations \cite{dogaru2023sphere, liu2020neural}, and more recently Gaussian splats \cite{kerbl20233d, huang20242d}. 

These recent advancements have dramatically expanded artistic expression into three-dimensional space, allowing us to freely change textures  \cite{chen2023text2tex, yeh2024texturedreamer, cao2023texfusion}, geometries  \cite{Yuan_2022_CVPR, binninger2024sens}, illumination \cite{rudnev2022nerfosr, gao2023relightable}, and introduce new objects into scenes without disrupting the environment \cite{gordon2023blended, shahbazi2024inserf}. This progress offers artists unprecedented creative freedom. Our research aims to extend these advanced 3D creation techniques into the microscopic world.

To explore this, we use images captured by a Scanning Electron Microscope (SEM). SEMs are used for the examination and analysis of nanoscale structures. An electron gun generates a beam that thoroughly scans the surface. As the beam interacts with the surface, it emits signals. The microscope's detectors capture these emitted electrons. The quantity of electrons detected from each point is then converted into corresponding pixel values, resulting in a high-resolution grayscale image that reveals the intricate surface structure. SEM images share similarities with optical images, exhibiting diffuse and specular reflectance and effects similar to optical shadowing. The fundamental distinction lies in the particle flow: in SEM imaging, the particles travel in the opposite direction compared to optical imaging.

We experiment with multi-view grayscale images of a pollen granule captured by tilting the sample while keeping the microscope fixed. However, tilting alters the incident angle between the electron beam and the surface, which causes the emitted electrons to vary across regions of the sample. This induces view-dependent variations in electron emission and scattering, which are perceived as illumination changes in the final SEM images.

Leveraging our captured grayscale dataset, we achieve novel view synthesis (NVS) via a precise 3D representation of the pollen modeled with 2D Gaussian Splatting (2DGS) \cite{huang20242d}. To address the aforementioned illumination variations, we apply an image specific affine color transformation (ACT) to the Gaussians, as proposed by \cite{darmon2024robust}. Moreover, based on the 3D representation, we further introduce colors into the scene, guided by artistic intuition. In addition to our grayscale dataset, we incorporate up to five color images created by a professional artist, Martin Oeggerli. These color images are then used for the colorization of the scene, by adapting ideas from \cite{zhang2023ref} to propagate the color information via pseudo-colors and semantic correspondences across views. We showcase the capabilities of our method, ArCSEM, by generating expressive colorized novel views of an SEM scene with artistic guidance.

The key contributions of our work are as follows:
\begin{itemize}
\item We obtain a precise and intricate 3D representation of a pollen captured by SEM, enabling novel view synthesis.
\item We achieve 3D colorization of the grayscale 3D scene using a limited number of manually colored images, enabling colored novel view synthesis.
\item We demonstrate the effectiveness of the proposed approach compared to previous methods through comprehensive experiments.
\end{itemize}

\begin{figure}[t]
    \centering
    \subfloat[][]{
        \centering
        \includegraphics[width=0.97\textwidth]{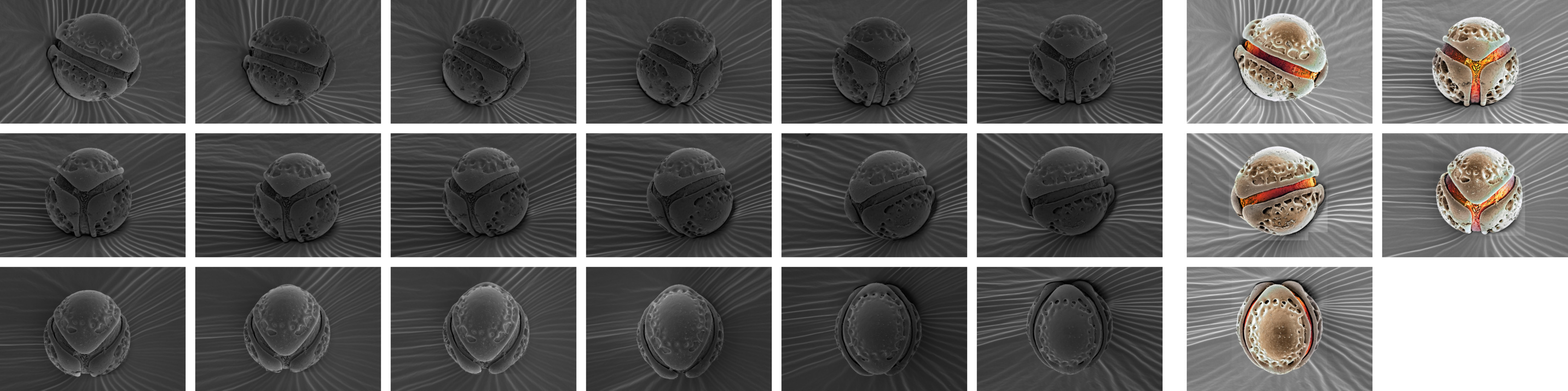}
        \label{fig:gray_dataset}
    }
    \subfloat[][]{
        \label{fig:color_dataset}
    }
    
    \vspace{-0.3cm}
    
    \hspace{0.28\textwidth}\text{(a) Grayscale} \hfill \text{(b) Color}\hspace{0.07\textwidth}
    
    \vspace{-0.1cm}
    \caption{Our dataset. (a) A subset of 18 out of 32 grayscale images, arranged left to right in the first two rows, and front to top in the bottom row. (b) All manually colored images shown in the following order: leftmost, center, rightmost, angled, and top view}
    \label{fig:dataset}
    \vspace{-1.5em}
\end{figure}

\section{Related work}

\subsubsection{Colorization.}
Colorization techniques can be broadly categorized into two main approaches: statistical methods and semantic methods. Statistical methods, pioneered by \cite{946629}, utilize color statistics of images, focusing on correcting overall color distributions between source and target images. While computationally efficient, these methods often lack semantic understanding, potentially leading to contextually inappropriate results. The advent of deep learning has transformed the field \cite{cheng2015deep}, making semantic methods more prevalent. These methods consider content-aware correspondences and leverage architectures based on Convolutional Neural Networks (CNNs) \cite{he2018deep, zhang2016colorful} or Transformers \cite{kumar2021colorization} to extract semantic features and produce more contextually coherent colorization. Beyond 2D images, 3D colorization methods have also been developed for meshes \cite{zhang2020deep}, point clouds \cite{gao2023scene, liu2019pccn}, and voxels \cite{yang2018learning}. The few works \cite{goytom2019nanoscale, venema2023colorizing} that consider the colorization of SEM images are limited to the 2D domain and work without specific color guidance. In contrast, we focus on artistic 3D colorization of SEM images, using specific color inputs, which allows artists to guide and control the process.

\subsubsection{3D SEM scenes.} Only one line of works \cite{zivanov2015multiview, zivanov2017reconstruction} has focused on 3D shape reconstruction of complex objects from SEM images, particularly of a cat flea. However, these approaches employ traditional photogrammetry and complex computer graphics techniques, which demand extensive mathematical calculations and laborious work. In contrast, our pipeline provides a much simpler method for representing 3D scenes, it is straightforward and requires no customization to model grayscale scenes.

\subsubsection{Appearance editing in NeRF and Gaussian splats.} Unlike 2D editing, 3D editing requires maintaining both geometric and appearance consistency across viewpoints while ensuring natural-looking edits. NVS techniques, such as Neural Radiance Fields (NeRF) \cite{mildenhall2021nerf} and its variants \cite{fridovich2022plenoxels, chen2022tensorf, barron2021mip}, have enabled the photorealistic rendering of arbitrary views. The versatility of these 3D representations enabled the development of various methods for appearance editing with diverse control modalities. Image-guided approaches \cite{zhang2022arf, nguyen2022snerf, zhang2023ref, huang2022stylizednerf, chiang2022stylizing, liu2023stylerf} leverage visual references to control the editing process. Text-based methods \cite{haque2023instruct, kim20233d, wang2022nerf, wang2023tsnerf, song2023efficient, dong2024vica, zhuang2023dreameditor} employ natural language descriptions to manipulate scene appearances. Other methods \cite{kuang2022palettenerf, radl2024laenerf, gong2023recolornerf, lee2023ice, mazzucchelli2024ireneinstantrecoloringneural} allow for manual color specification or tool-based editing.  
Recently, Gaussian Splatting \cite{kerbl20233d, huang20242d} has emerged as a breakthrough technique, representing scenes explicitly with numerous 3D Gaussians primitives. 
Providing real-time rendering and competitive quality, the framework has already enabled several editing techniques \cite{liu2024stylegaussian, zhang2024stylizedgscontrollablestylization3d, chen2023gaussianeditor}. Our proposed method belongs to the latter line of works, but, in contrast to existing approaches that mainly focus on stylization or color replacement using extracted color palettes, we build on Ref-NPR \cite{zhang2023ref} which is more suitable for our grayscale-to-color setting.

\section{Method}
\begin{figure}[tb]
  \centering
  \includegraphics[width=\textwidth]{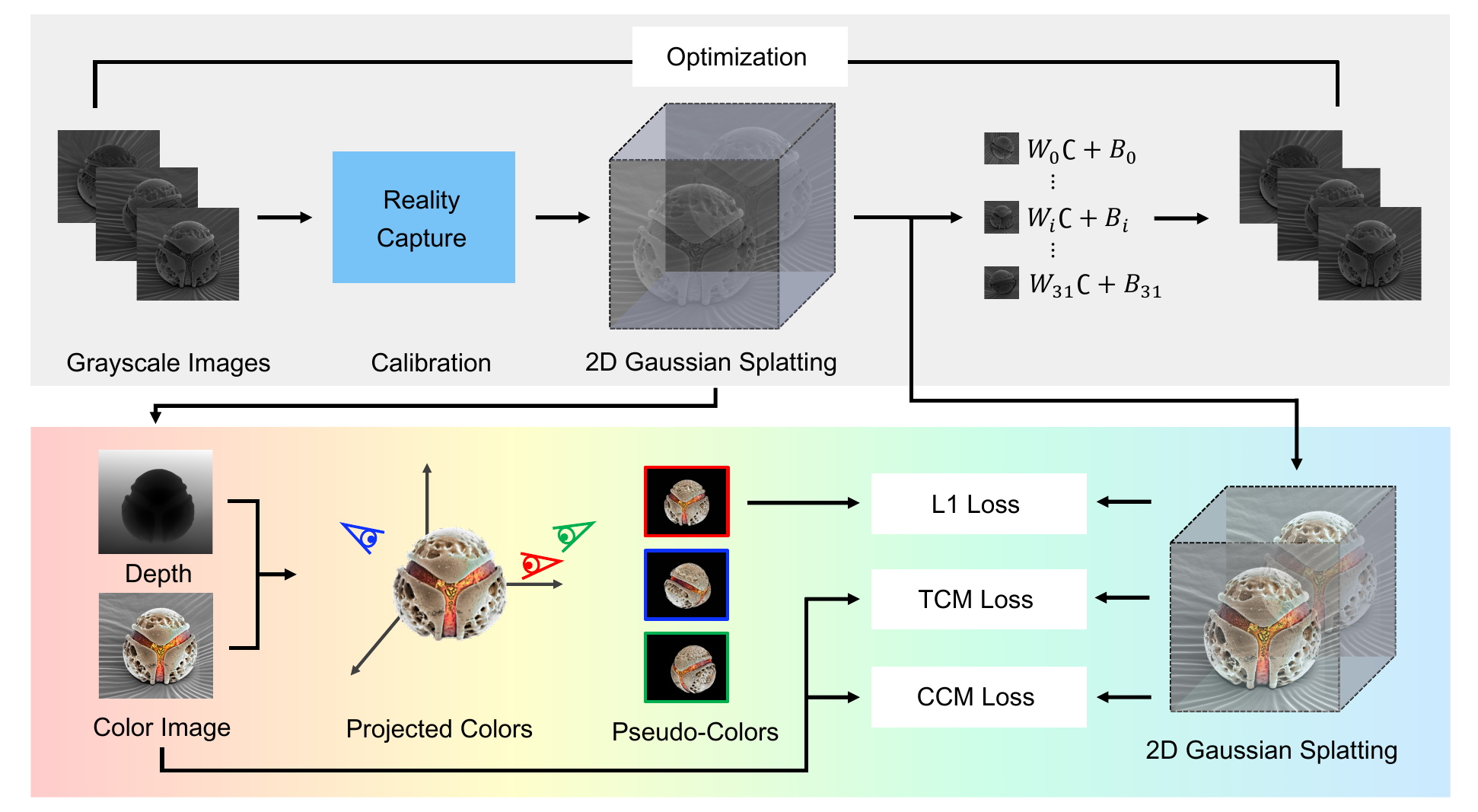}
  \caption{Overview of our two-stage approach: (a) \textbf{Grayscale training}: 
  We fit 2DGS \cite{huang20242d} with an image-specific affine color transformation to the grayscale images calibrated with RealityCapture \cite{realitycapture}. 
  (b) \textbf{Colorization}: 2DGS depth maps are used to project colors from limited manually colorized images into 3D space as pseudo-colors. Together with the input color views, the pseudo-colors guide the colorization of the grayscale model via L1, TCM, and CCM loss functions.}
  \label{fig:overview}
  \vspace{-1em}
\end{figure}

Our proposed method, illustrated in Fig.~\ref{fig:overview}, employs a two-stage training process: grayscale 3D scene optimization and colorization. We start with a grayscale scene representation by fitting 2DGS on the SEM image dataset calibrated with RealityCapture \cite{realitycapture}. To handle varying illumination, we apply an affine color transformation to the decoded color, using image-specific weights and biases. In the second stage, we use the grayscale model to generate depth maps and project the artist-provided colors into 3D space. For views without color data, we use a nearest-neighbor search to obtain pseudo-colors. Finally, we fine-tune the initial grayscale model using the color images and the computed pseudo-colors by keeping the geometry fixed and optimizing the spherical harmonics coefficients for all degrees using losses inspired from Ref-NPR \cite{zhang2023ref}.

\subsection{Grayscale Training}
\subsubsection{Calibration.}
Our grayscale dataset is acquired using SEM. Although SEM utilizes parallel electron beams, resulting in orthographic projection, we approximate it using perspective projection to ensure compatibility with existing NVS methods. Under this setting, an exceptionally large focal length of approximately 50,000 pixels is estimated, with the cameras positioned very far from the scene content. To this end, we experimented with COLMAP \cite{schonberger2016structure, schonberger2016pixelwise} and more advanced feature extraction and matching methods, including SuperGlue \cite{sarlin2020superglue} and LoFTR \cite{sun2021loftr}, but found them to struggle with the peculiar images produced by SEM. Instead, we obtained satisfying results using RealityCapture \cite{realitycapture} with shared intrinsic parameters across all images, while allowing for image-specific distortion coefficients. This approach yielded the most accurate and highly precise calibration for our dataset.

\subsubsection{Backbone model.}
While the original method we built upon, Ref-NPR \cite{zhang2023ref}, utilizes Plenoxels \cite{fridovich2022plenoxels}, we found this backbone to be inappropriate for our setup, often resulting in visual artifacts (see Fig.~\ref{fig:gray_closeup}). 
Alternatively, we considered the more recent 3D Gaussian Splatting \cite{kerbl20233d} method which represents the scene using 3D Gaussian primitives, but encountered similar issues including visible floaters caused by poor geometry fitting. 
Lastly, the successor method 2DGS \cite{huang20242d}, which uses 2D oriented Gaussian disks to represent the scene, succeeded in modeling the SEM images, rendering accurate depth maps and grayscale images. 

\subsubsection{View-specific effects.}
To accommodate varying illumination conditions, we employ an affine color transformation (ACT) as proposed by \cite{darmon2024robust}. Each Gaussian holds spherical harmonics coefficients, which are subsequently converted to output intensity values. Prior to rasterization, we apply an image-dependent transformation on the decoded illumination, $\mathbf{L}$. During the subsequent colorization stage, we found that applying this affine transformation led to a degradation in output quality, so we omit it in the second stage. The transformation uses three weights $\mathbf{W} = \{w_1, w_2, w_3\}$ and biases $\mathbf{b} = \{b_1, b_2, b_3\}$ and is defined as $ \mathbf{L}' = \mathbf{W} \cdot \mathbf{L} + \mathbf{b} $. When rendering novel views, we average the weights and biases of the training views, resulting in plausible and consistent illumination, as can be seen in Fig.~\ref{fig:gray_novel}.

\subsection{Colorization}
Our colorization method is based on Ref-NPR \cite{zhang2023ref}, which was introduced for 3D stylization. We made several key modifications and enhancements to adapt it to our dataset and colorization needs. This section details our approach, highlighting the differences and improvements. We rely on three components: pseudo-color supervision for views lacking color information, a Template-based Correspondence Module for propagating colors via the grayscale feature space, and a Coarse Color-Matching Loss to ensure global color consistency.

\subsubsection{Pseudo-color supervision.}
For color transfer, we first utilize the depth information of the grayscale model to unproject the input pixels of the colored views into the 3D space. Then, for each pixel in the other views we compute a pseudo-color as the color of the closest colored point to its unprojected location. This color is then used as a supervision signal via the loss defined in Eq.~\ref{eq:pseudo_color}, where \( \hat{\mathbf{C}}_{\text{pc}} \) denotes the pseudo-color, and \( \hat{\mathbf{C}}_{\hat{x}} \) refers to the color rendered by the model. If there is no colored point within a given radius, we exclude the respective pixel from the loss calculation. 

\begin{equation}
\mathcal{L}_{\text{pseudo-color}} = \frac{1}{N_{\text{pc}}} \left\| \hat{\mathbf{C}}_{\text{pc}} - \hat{\mathbf{C}}_{\hat{x}} \right\|_1.
\label{eq:pseudo_color}
\end{equation}

Ref-NPR employs Reference Ray Registration with a grid system, where colors are assigned to grids and a single color is selected for each pixel. However, this approach does not consider the distance between points during the selection phase; it only filters out pixels that exceed a threshold in the final image, making it difficult to create precise pseudo-colors and potentially resulting in artifacts in the final output.

Our approach is designed to accommodate high-resolution datasets without encountering memory constraints. The grid-based method in Ref-NPR becomes computationally infeasible for our data, as the higher number of required grids to match our cinematic resolution would exhaust available memory resources.

While Ref-NPR also considers the cosine similarity of ray directions in the final image, we found that this factor had minimal impact on quality for our dataset. Therefore, we simplified our approach by concentrating exclusively on spatial proximity.

Pseudo-colors for the background regions often introduce noise and inconsistencies across images, as the number of background pixels varies significantly between images, potentially skewing the loss calculation in Eq.~\ref{eq:pseudo_color}. To address these issues, we employ Segment Anything Model \cite{kirillov2023segment} to generate precise masks of the pollen granule. This segmentation allows us to effectively extract only the pollen and eliminate the interference of the background elements.

\subsubsection{Template-based Correspondence Module.}
We employ TCM proposed by Ref-NPR as a loss function to propagate colors to the areas that do no have a ground truth color assigned in the colorized input views, by using matches in the feature space of the grayscale images.  This loss minimizes the cosine distance between the features \( F_{\hat{I_g}} \) of the rendered color image and a constructed guidance feature \( \hat{F}_{I_g} \) of the view \( I_g \). 
The grayscale image \( I_g \) is fed into a VGG network \cite{7486599} to extract the feature map \( F_{I_g} \). The feature maps of the reference colorized images $S_k$ and their grayscale version $I_k$ are extracted as $F_{S_k}$ and $F_{I_k}$ respectively. For each location $(i,j)$ in the guidance feature map \( \hat{F}_{I_g}^{(i, j)} \), we consider the grayscale feature \( F_{I_g}^{(i, j)} \) and search for the nearest grayscale feature across reference views \( F_{I_k}^{(i^*, j^*)} \) and take the corresponding feature of the colorized image \( F_{S_k}^{(i^*, j^*)} \) , as defined in Eq.~\ref{eq:tcm_search}. 
Further details on TCM can be found in \cite{zhang2023ref}.

\begin{equation}
\vspace{-1.0em}
\hat{F}_{I_g}^{(i,j)} = F_{S_k}^{(i^*, j^*)}, \quad \text{where} \quad (i^*, j^*), k = \arg \min_{(i', j'), k'} \text{dist}\left(F_{I_g}^{(i,j)}, F_{I_{k'}}^{(i', j')}\right),
\label{eq:tcm_search}
\end{equation}

\begin{equation}
\mathcal{L}_{\text{TCM}} = \text{dist}(F_{\hat{I_g}}, \hat{F}_{I_g})
\label{eq:tcm_loss}
\vspace{-1em}
\end{equation}

\subsubsection{Coarse Color-Matching Loss.}
Although TCM helps estimate color for occluded regions, it can result in global color inconsistencies and mismatches. To address this limitation, we also consider a coarse color-matching loss \cite{zhang2023ref} that operates at the patch level to minimize color differences, as defined in Eq. \ref{eq:coarse_color}. Using the index $(i^*,j^*)$ obtained in Eq. \ref{eq:tcm_search}, let $\bar{C}$ denote the average color of a patch, and $C_{S_k}$ and $C_{I_g}$ refer to the patches in the input color image and rendered image respectively. See \cite{zhang2023ref} for details.

\begin{equation}
\mathcal{L}_{\text{coarse-color}} = \frac{1}{N} \sum_{i,j} \| \bar{C}_{I_g}^{(i,j)} - \bar{C}_{S_k}^{(i^*,j^*)} \|_2^2.
\label{eq:coarse_color}
\vspace{-1em}
\end{equation}

\subsubsection{Optimization.} For views with available colorized image, we directly optimize the L1 loss between rendered and input images. For the other views, our final loss function is defined in Eq. \ref{eq:loss_func}. $\lambda$s are respective weights for each loss term. 

\begin{equation}
\mathcal{L} = \lambda_{\text{pc}} \mathcal{L}_{\text{pseudo-color}} + \lambda_{\text{TCM}} \mathcal{L}_{\text{TCM}} + \lambda_{\text{cc}} \mathcal{L}_{\text{coarse-color}}
\label{eq:loss_func}
\end{equation}

\section{Experiments}

\subsection{Dataset}
Our dataset comprises 32 high-resolution ($3072 \!\times\! 2048$) SEM images of a pollen, captured along two primary axes. The horizontal axis consists of 20 images spanning from left to right, providing a comprehensive lateral view, while the vertical axis includes 12 images, with the camera moving in an arc from the frontal view to the top view of the pollen. We illustrate 18 of these grayscale images in Fig.~\ref{fig:gray_dataset}, and use the entire set of 32 images as our training dataset. Fig.~\ref{fig:color_dataset} shows colorized versions of 5 of the grayscale images, manually colorized by an artist known for his work on microscopic subjects, Martin Oeggerli.

\subsection{Implementation details}
Plenoxels \cite{fridovich2022plenoxels} uses an initial grid resolution of $128 \!\times\! 128 \!\times\! 128$, which is increased up to $304 \!\times\! 304 \!\times\! 128$. Larger grid resolution led to training collapse, likely due to variations in scene illumination. The model was trained for 38,400 iterations at each grid resolution, with a total of 115,200 iterations. Using Plenoxels trained on grayscale, we train Ref-NPR for 10 epochs in the colorization phase. To address the computational and memory demands of the Template-based Correspondence Module (TCM) and coarse color-matching (CCM) loss, we compute them on images scaled down by a factor of 4. The grayscale 3D Gaussian Splatting (3DGS) \cite{kerbl20233d} and 2D Gaussian Splatting (2DGS) \cite{huang20242d} models are trained for 60,000 epochs, with 20,000 additional epochs for the colorization process. 
We initialize the affine color transformation (ACT) by adding small random perturbations to the identity transformation (weights $w_i = 1$ and biases $b_i = 0$, for $i = 1, 2, 3$) and optimize the parameters using a learning rate of 0.0001. 

\subsection{Grayscale}

\begin{figure}[tb]
    \vspace{-0.1cm}
    \begin{subfigure}{\textwidth}
        \centering
        \includegraphics[width=\textwidth]{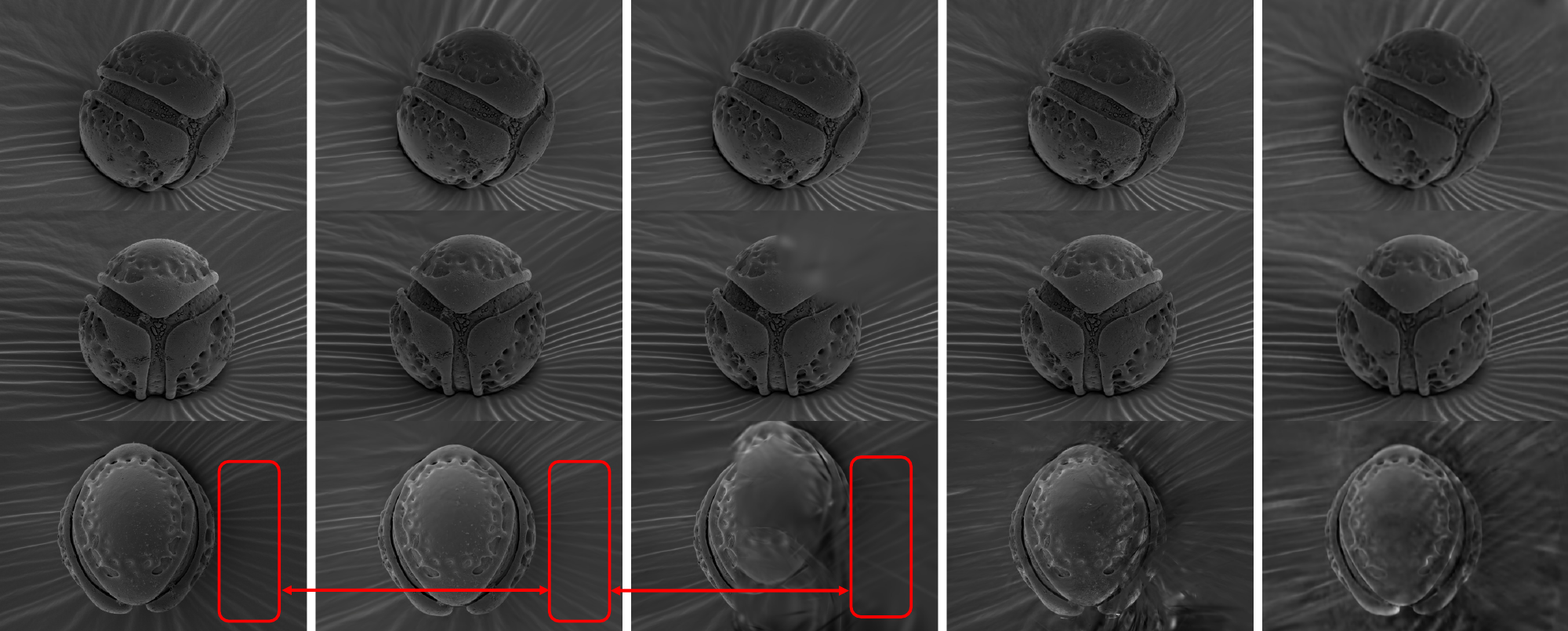}
    \end{subfigure}

    \begin{tabularx}{\textwidth}{XXXXX}
        \centering GT &
        \centering Ours & \centering 2DGS \cite{huang20242d} & \centering 3DGS \cite{kerbl20233d} & \centering Plenoxels \cite{fridovich2022plenoxels} 
    \end{tabularx}   
    \vspace{-0.5cm}

    \caption{Grayscale novel views. The first two rows show novel views in the lateral trajectory and the bottom row indicates the view from the top. All models are trained at $3072 \!\times\! 2048$. Red squares highlight areas with illumination differences. ACT in Ours effectively normalizes the differences across views.}
    \vspace{-0.5cm}
    \label{fig:gray_novel}
\end{figure}

\begin{figure}
  \centering
  \includegraphics[width=\textwidth]{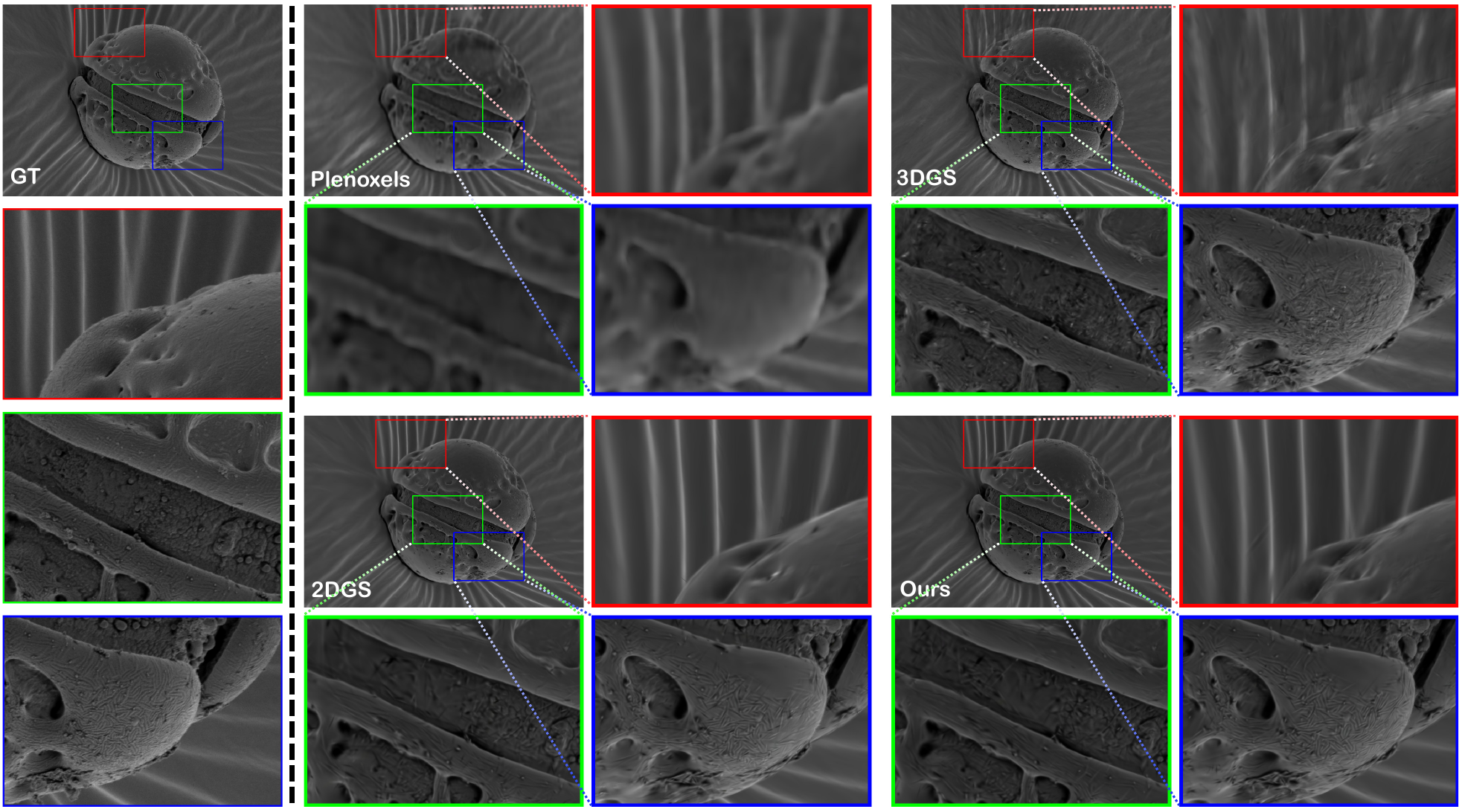}
  \caption{Grayscale novel views with closeups. The left side displays the nearest Ground Truth (GT) image along with its closeups. This novel view is precisely between two adjacent GTs. The right side shows the generated novel views and their corresponding closeups. Our method (2DGS+ACT) exhibits superior quality in synthesising novel views.}
  \label{fig:gray_closeup}
  \vspace{-2em}
\end{figure}

We qualitatively compare the backbones on grayscale novel view synthesis and quantitatively on rendering the training views by evaluating common image quality metrics: PSNR, SSIM \cite{wang2004image}, and LPIPS \cite{zhang2018unreasonable}. All models were trained at the full $3072 \!\times\! 2048$ resolution. Although Plenoxel's grid resolution was insufficient to fully represent this high resolution, we prioritized consistent training conditions to ensure a fair comparison of the best possible rendered quality across all methods especially since rendering at cinematic resolution is our goal.

The rendered novel views are presented in Fig.~\ref{fig:gray_novel}, with detailed close-ups shown in Fig.~\ref{fig:gray_closeup}. We observe in Fig.~\ref{fig:gray_novel} that our method demonstrates the ability to generate high-quality novel views with notable fidelity compared to other methods. ACT enables 2DGS to achieve scene representation without the prominent floaters that often appear in other approaches, especially in the view from the top. Additionally, in the bottom view (red squares), our method produces a slightly brighter image compared to the ground truth and 2DGS. This observation indicates that ACT effectively normalizes illumination across different viewpoints, resulting in more consistent lighting conditions, as well as improved rendering quality. As depicted in Fig.~\ref{fig:gray_closeup}, while Plenoxels lacks fine details, the other methods model the scene with high precision. Still, 3DGS exhibits various artifacts, which are not observed in either 2DGS or 2DGS+ACT.

\begin{figure}
    \vspace{-0.8em}
    \centering
    \begin{subfigure}{\textwidth}
        \centering
        \includegraphics[width=\textwidth]{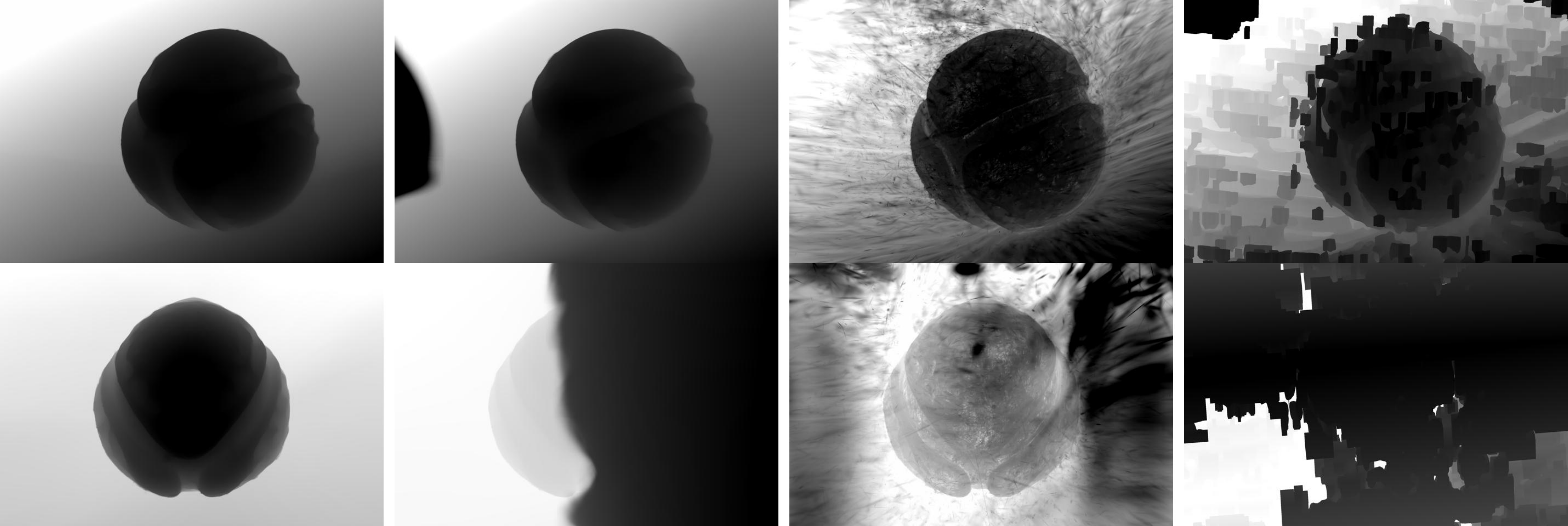}
    \end{subfigure}

    \vspace{0.05cm}
    \begin{tabularx}{\textwidth}{XXXX}
        \centering Ours & \centering 2DGS \cite{huang20242d} & \centering 3DGS \cite{kerbl20233d} & \centering Plenoxels \cite{fridovich2022plenoxels} 
    \end{tabularx}
    \vspace{-0.5cm}
    \caption{Predicted depth maps of novel views. The view in the first row is sampled from the horizontal trajectory, while the view in the second is sampled from the vertical trajectory. All other methods failed to predict depths in the vertical trajectory. The depth map accuracy is essential for correctly projecting colors into the 3D space.}
    \vspace{-1.8em}
    \label{fig:depth_maps}
\end{figure}

The rendered depth maps presented in Fig.~\ref{fig:depth_maps} are consistent with the grayscale predictions: all other approaches face challenges in accurately estimating depths from top viewpoints. This is due to significant illumination variations in the top views (see the bottom row of Fig.~\ref{fig:gray_dataset}).
Notably, 3DGS exhibits significant artifacts, not only in top views, but also in lateral perspectives. In contrast, our approach generates remarkably clean and consistent depth maps for the entire scene, demonstrating superior performance across various viewpoints.

\begin{table}
\vspace{-0.5em}
  \centering
  \begin{tabular}{@{\hskip 10pt}l@{\hskip 20pt}l@{\hskip 20pt}l@{\hskip 20pt}l@{\hskip 10pt}}
    \toprule
    Model & PSNR ($\uparrow$) & SSIM ($\uparrow$) & LPIPS ($\downarrow$) \\
    \midrule
    Plenoxels \cite{fridovich2022plenoxels} & 30.94 & 0.855 & 0.556 \\
    3DGS  \cite{kerbl20233d}& \textbf{37.50} & \textbf{0.902} & \textbf{0.461} \\
    2DGS \cite{huang20242d}&  35.25 & 0.867 & 0.511 \\
    2DGS + ACT (Ours) & 36.32 & 0.890 & 0.489 \\
    \bottomrule
  \end{tabular}
  \vspace{0.5em}
  \caption{Evaluation of rendered training images. 3DGS and our method achieve the best quantitative scores for reproducing the training views, demonstrating the ability of modelling SEM images.}
  \label{tab:gray_metrics}
  \vspace{-0.5em}
\end{table}

Although the quantitative evaluation presented in Tab.~\ref{tab:gray_metrics} indicates that 3DGS outperforms in terms of rendering quality of the training views, 2DGS actually produces superior results in terms of perceived qualitative novel view quality (see Fig.~\ref{fig:gray_closeup}). Moreover, the integration of ACT improves the results of 2DGS across all considered metrics. We do not quantitatively evaluate the methods for NVS in order to avoid further reducing the already limited training set.

\subsection{Colorization}
\subsubsection{Qualitative evaluation.}
\begin{figure}
  \centering
  \vspace{-1em}
  \includegraphics[width=\textwidth]{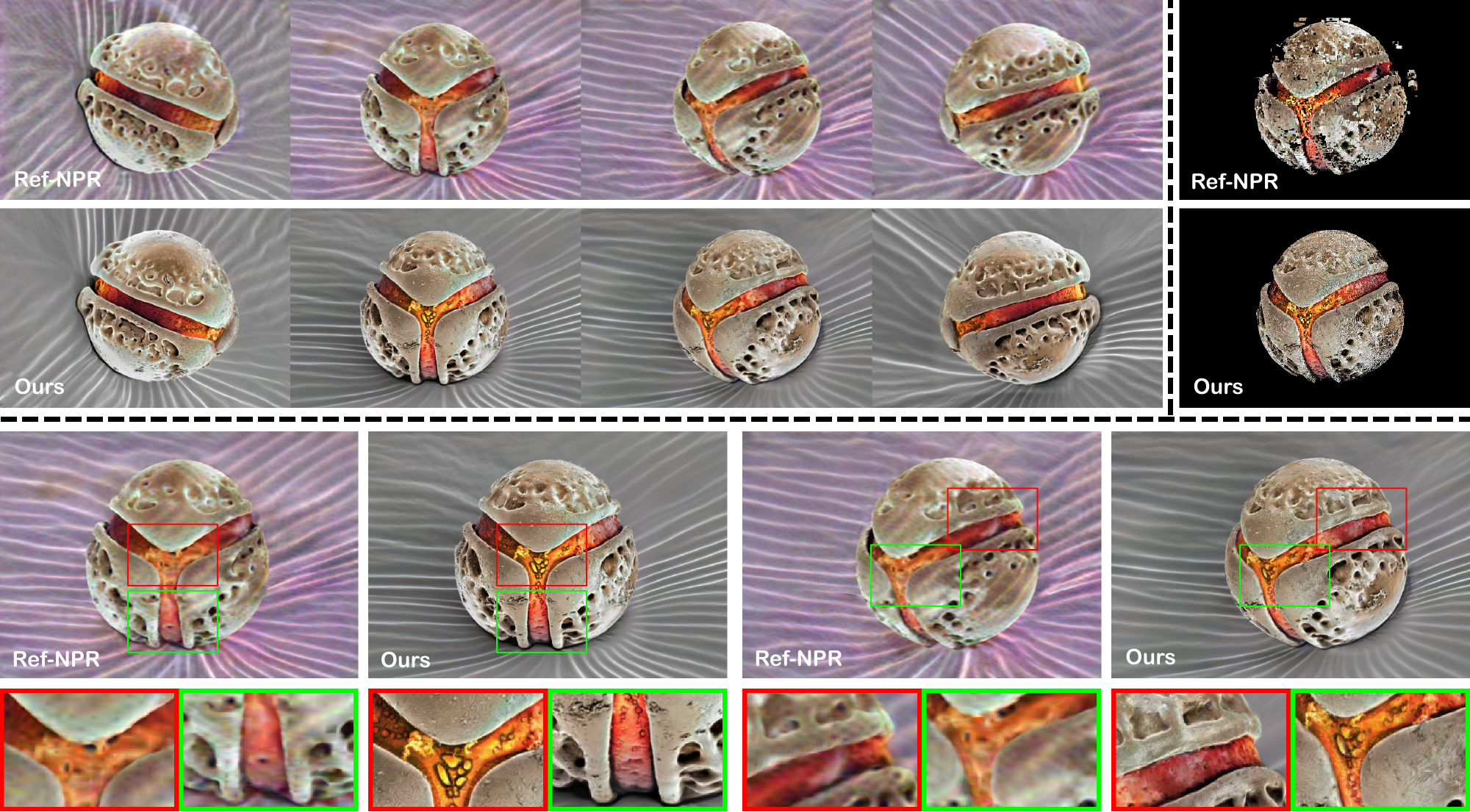}
  \caption{Qualitative comparison between Ref-NPR \cite{zhang2023ref} and our method, both trained on $768 \!\times\! 512$ resolution images with five colored images. Top left: Synthesized novel views. Top right: Pseudo-colors. Bottom: closeups of novel views.}
  \label{fig:refnpr_comparison}
  \vspace{-2.0em}
\end{figure}

We compare our method against Ref-NPR \cite{zhang2023ref} for colorization using five color images as input and present the results in Fig.~\ref{fig:refnpr_comparison}. For fairness, both methods were trained at a lower resolution ($768 \!\times\! 512$), due to Plenoxel's limitation on grid resolution. Note that, at this grid resolution, the grayscale rendering is of similar quality at low and high image resolution, as depicted in Fig.~\ref{fig:gray_novel}. The rendered views by Ref-NPR exhibit noticeable purple artifacts in the background. To address this, the method uses CCM loss. However, our experiments revealed that excessive reliance on this loss affected the fine details. Showcased here is the best result achieved in our experiments with Ref-NPR. Our method, in contrast, successfully colorizes even the very fine details, particularly in the complex geometry and appearance of the pollen's center. Regarding pseudo-colors, our method generates nearly perfect pseudo-colors, whereas Ref-NPR's grid-based color selection method results in voxel-like artifacts. Although our method also shows some inconsistencies at the pseudo-color level, the final results exhibit a smooth transition on the surface color.

\subsubsection{High-resolution colorization.}
To obtain the final result, we trained our model with five color images at the original resolution ($3072 \!\times\! 2048$). The generated novel views and corresponding close-ups are shown in Fig.~\ref{fig:color_closeup}. Our method accurately captures fine details such as black outlines of small circular protrusions at the center and the gradual transition of red hues from center to periphery. The boundary of the pollen is clear, and color changes on the gray surface are also captured. Overall, our method achieved uniform and high-quality colorization across the entire scene.

\subsubsection{Different number of color inputs.}
\begin{figure}
  \centering
  \includegraphics[width=\textwidth]{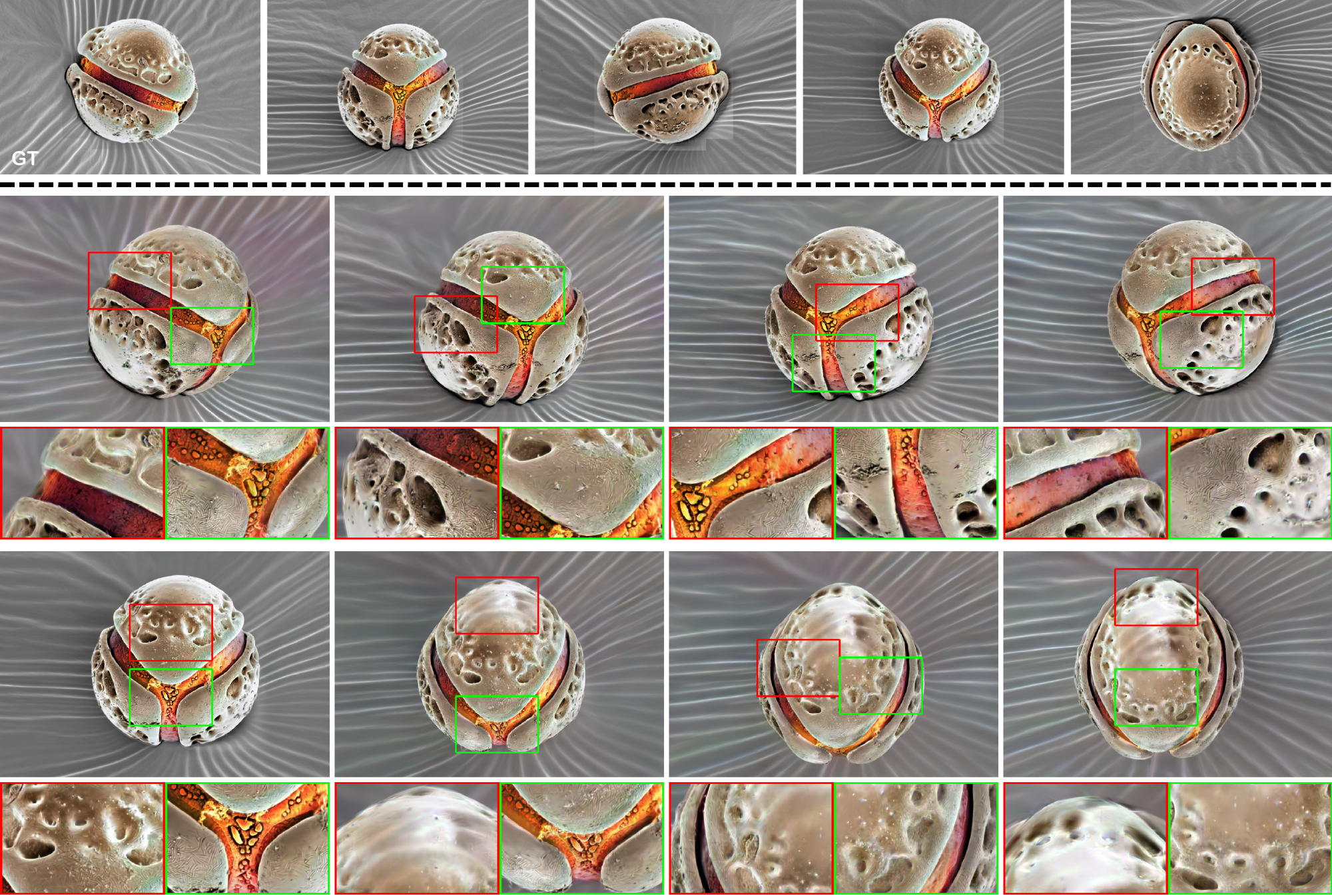}
  \caption{Novel views and closeups generated by our method. Top: All input color images. Bottom: Synthesized novel views at $3072 \!\times\! 2048$ resolution, with corresponding closeups.}
  \label{fig:color_closeup}
  \vspace{-1.5em}
\end{figure}

\begin{figure}[tb]
  \centering
  \begin{subfigure}{\textwidth}
    \centering
    \includegraphics[width=0.98\textwidth]{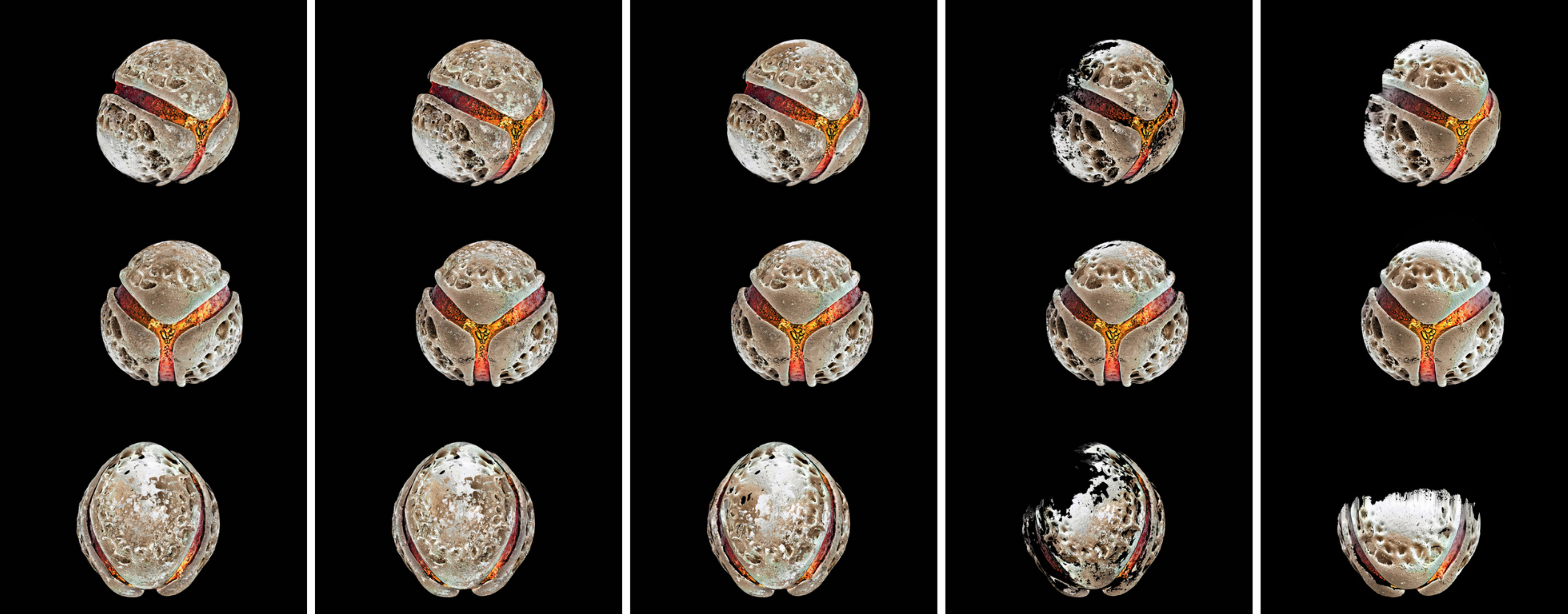}
  \end{subfigure}
  
  \vspace{0.5em}

  \begin{subfigure}{\textwidth}
    \centering
    \includegraphics[width=0.98\textwidth]{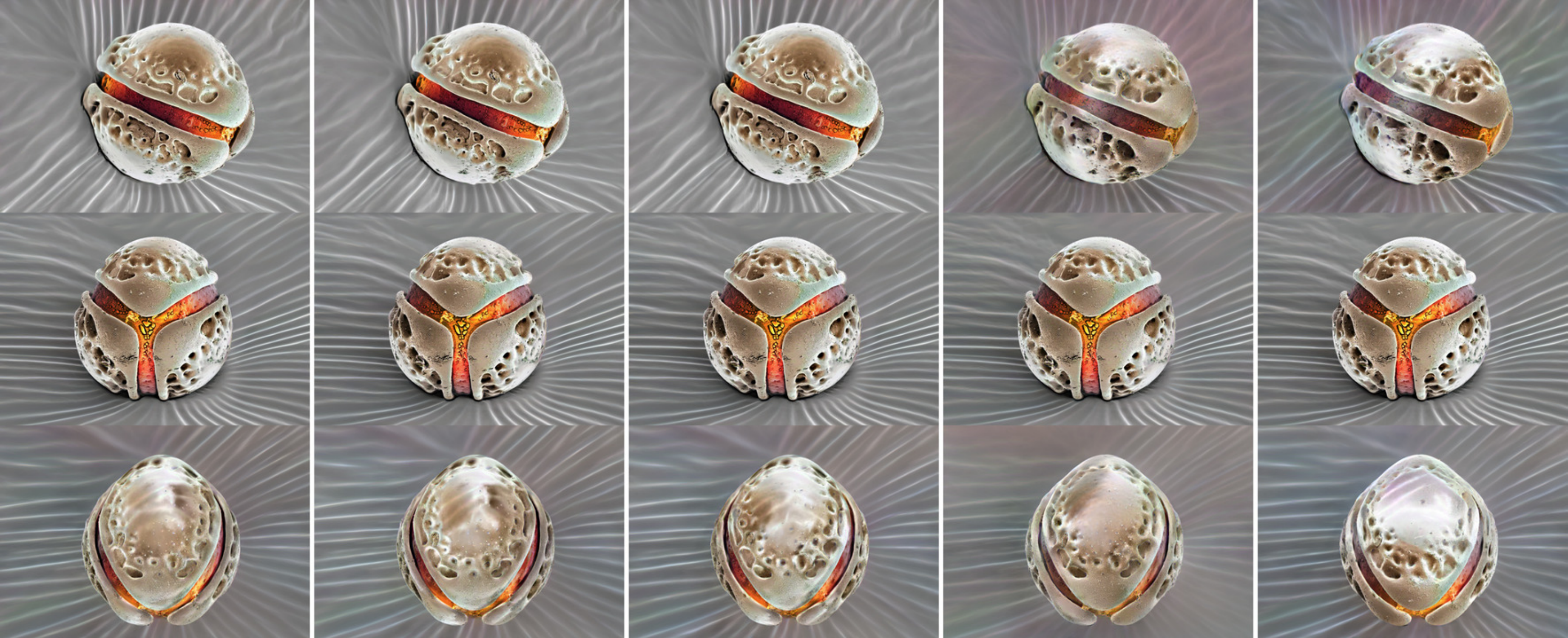}

    \begin{tabularx}{\textwidth}{XXXXX}
        \centering 5 images & \centering 4 images & \centering 3 images & \centering 2 images & \centering 1 images
    \end{tabularx}
    
    \label{fig:different_colors_res}
  \end{subfigure}
  
  \caption{Novel views with varying amounts of manually colored input images. Whilst for a limited range of viewpoints we reach excellent results even from a single colored image, more images improve the overall colorization quality visible (especially when animated).}
  \label{fig:diff_colors}
  \vspace{-1em}
\end{figure}

Our final results utilize five color images, but we also investigated the impact of using fewer artist colorized images to guide the colorization of the scene. We examined scenarios with one, two, three, and four color images. The four-image case omits the top view (rightmost in Fig.~\ref{fig:color_closeup}). The three-image case uses only lateral views. The two-image case employs the frontal and rightmost lateral views. The one-image case relies only on the frontal view. Fig.~\ref{fig:diff_colors} shows the pseudo-colors and the corresponding colorization results with varying numbers of color inputs. In the single-image case, which only uses the frontal view, it tends to produce darker red on the sides. Strong reflections appear on the sides and top. This effect originates from the pseudo-colors, as seen in the first row of the single image case. Nevertheless, even with just one image, our method achieves reasonably accurate colorization. In the two-image case, which lacks the leftmost view, the predicted colors are also darker on the sides, though not as strong as in the single-image case. In the three- and four-color image cases, which lack the top views, slight color inconsistencies are noticeable on the top compared to the five-image case. As the area covered by color inputs increases, the colorization accuracy improves.

\subsection{Ablation study}
To evaluate the contribution of each component in our proposed model, we conducted an ablation study. We compared the full model against versions without TCM, without CCM, without ACT, and without all components. 

\begin{figure}[t]
  \centering
  \includegraphics[width=0.98\textwidth]{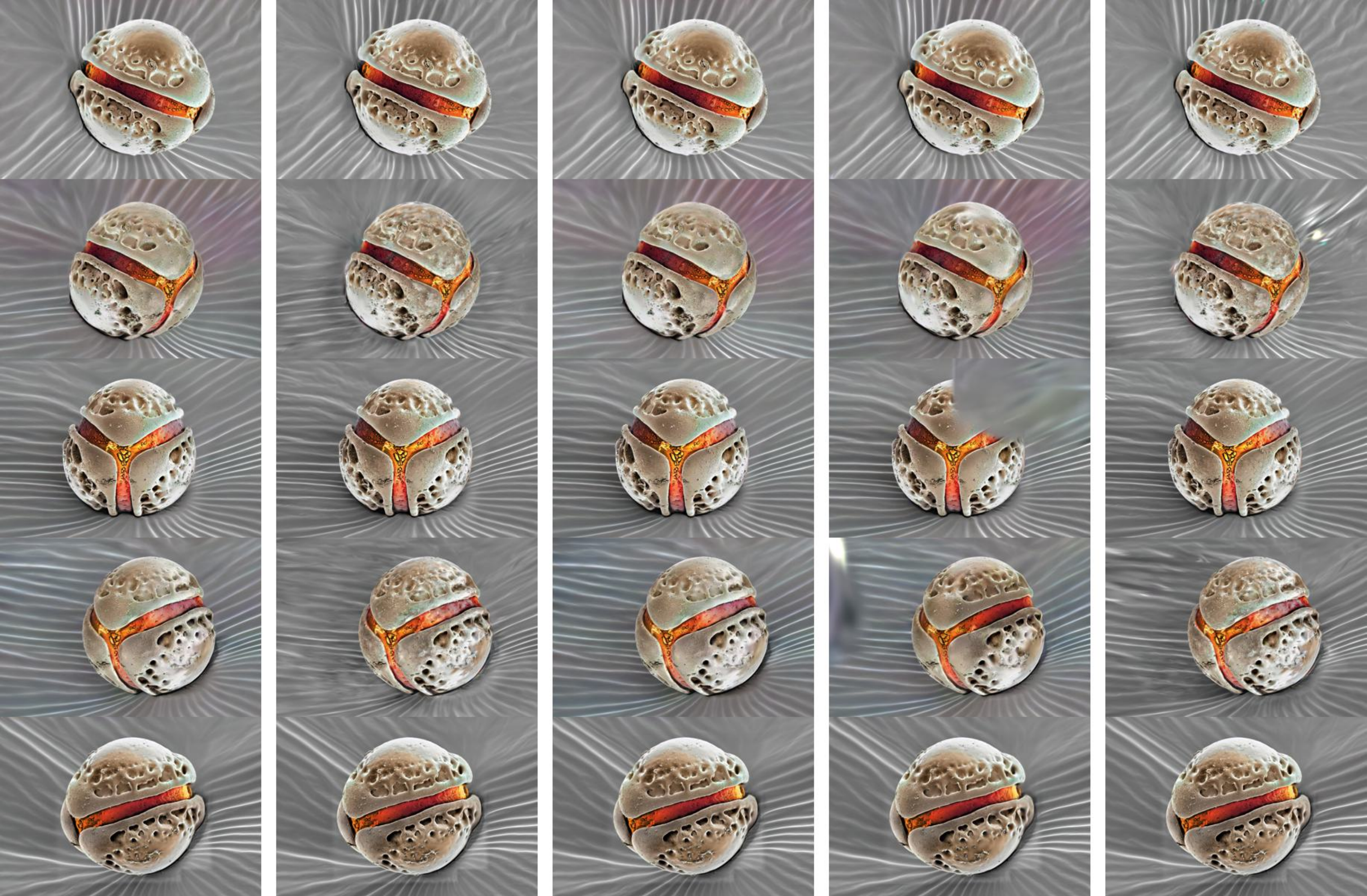}
    \begin{tabularx}{\textwidth}{XXXXX}
        \centering Full model & \centering w/o TCM & \centering w/o CCM & \centering \text{w/o ACT} & \centering \text{w/o all}
    \end{tabularx}
    
  \caption{
    Ablation studies. 
    This figure illustrates the impact of different model components on colorization quality. 
    From left to right: Our full model, without Template-based Correspondence Module  (TCM), 
    without coarse color-matching (CCM) loss, without affine color transformation (ACT), 
    without all modules. 
  }
  \vspace{-1em}
  \label{fig:ablation_study}
\end{figure}

Fig.~\ref{fig:ablation_study} illustrates the generated novel views from this study. The case (w/o TCM) shows noticeable color inconsistencies on the surface, likely due to variations in the pseudo-colors. This highlights the effectiveness of TCM in estimating colors, although it introduces subtle background color shifts, with slight reddish (second row) or bluish tints (fourth row) in the Full model case. Still, the CCM ablation shows the role of the loss in reducing global color inconsistencies, as these background color artifacts are mitigated in the full model. The case without ACT leads to more pronounced artifacts (third row), originating from the grayscale representation (see Fig.~\ref{fig:gray_novel}). Additionally, reddish background color shifts are also seen in this case, and the color on the sides (fourth row) is much darker than in other cases. This phenomenon is likely due to the presence of these floaters, which adversely impacted the effectiveness of CCM, leading to its failure. Moreover, compared to the case without any components, the color prediction in our full model performs remarkably well. Consequently, each component contributes to our final results.

\section{Limitations}
Our method requires substantial processing time on high-end hardware. Generating pseudo-colors takes approximately 5 hours on an NVIDIA A100 GPU due to the need for each pixel to identify the nearest color from among all pixels across the five color images of pollen. Additionally, the colorization process itself takes around 3 hours. This time requirements are induced by TCM and the CCM loss.

While our method produces high-quality results, some challenges remain. As illustrated in Fig.~\ref{fig:gray_novel}, accurately modeling very fine surface patterns in the grayscale stage has proven challenging. Moreover, as depicted in Fig.~\ref{fig:gray_novel}, artifacts persist in the colorization results. These include a subtle reddish tint in the background, minor green discoloration near the central raised area and along the sides, line-shaped artifacts in the red regions on both sides, and a blurred area in the apical region.

As illustrated in Fig.~\ref{fig:diff_colors}, the current single-view case is limited to the colors visible in the image. However, this limitation could potentially be overcome by employing segmentation or feature-based approaches, or even a diffusion-based method to estimate the colors of the unseen parts of the object.

\section{Conclusion}
We achieved cinematic colorization of pollen images captured by a Scanning Electron Microscope. Our approach, which incorporates color projection onto 3D space, affine color transformation, a Template-based Correspondence Module, and a Coarse Color-Matching loss, demonstrated superior performance on our dataset compared to existing methods. We effectively validated the necessity and efficacy of each component in achieving our results. From an artistic perspective, we are confident in the value of our work; introducing color to a monochrome realm offers a visually arresting and mesmerizing experience. Moreover, our method particularly enables us to reduce the number of viewpoints artists need to color manually. By eliminating the manual annotations through our novel view synthesis process, our approach not only enhances efficiency but also opens new creative possibilities for artists.

\section*{Acknowledgments}
We would like to thank Maximilian Weiherer for valuable discussions and support with the camera calibration.
Andreea Dogaru was funded by the German Federal Ministry of Education and Research (BMBF), FKZ: 01IS22082 (IRRW). The authors are responsible for the content of this publication.
The authors gratefully acknowledge the scientific support and HPC resources provided by the Erlangen National High Performance Computing Center (NHR@FAU) of the Friedrich-Alexander-Universität Erlangen-Nürnberg (FAU) under the NHR project b112dc IRRW. NHR funding is provided by federal and Bavarian state authorities. NHR@FAU hardware is partially funded by the German Research Foundation (DFG) – 440719683. This project was supported by the special fund for scientific works at the Friedrich-Alexander-Universität Erlangen-Nürnberg.

\bibliographystyle{splncs04}
\bibliography{main}
\end{document}